\pdfoutput=1
\documentclass[11pt]{article}

\usepackage[]{acl}
\usepackage{times}
\usepackage{latexsym}
\usepackage{enumitem}

\usepackage{subfig}

\usepackage[T1]{fontenc}
\usepackage[utf8]{inputenc}
\usepackage{graphicx}
\usepackage{amsmath}
\usepackage{algorithm}
\usepackage{algpseudocode}
\usepackage{microtype}

\title{CaPE: Contrastive Parameter Ensembling for Reducing Hallucination in Abstractive Summarization}

\author{Prafulla Kumar Choubey$^1$ \quad Alexander R. Fabbri$^1$ \quad Jesse Vig$^1$ \\
 {\bf Chien-Sheng Wu$^1$ \quad \stepcounter{footnote}Wenhao Liu$^{2}$~\thanks{{ } work was done at Salesforce AI Research.} \quad Nazneen Rajani$^{3}$~\footnotemark[2]} \\
$^1$Salesforce AI Research, $^2$Faire.com, $^3$Hugging Face \\
\texttt{\{pchoubey, afabbri, jvig, wu.jason\}@salesforce.com} \\
\texttt{wenhao@faire.com, nazneen@hf.co} \\
}
\begin{document}
\maketitle
\begin{abstract}

% \jw{
Hallucination is a known issue for neural abstractive summarization models. Recent work suggests that the degree of hallucination may depend on errors in the training data. In this work, we propose a new method called Contrastive Parameter Ensembling (CaPE) to use training data more effectively, utilizing variations in noise in training samples to reduce hallucination. We first select clean and noisy subsets from the training data using different automatic factual metrics. Then, we fine-tune a base summarization model, which is trained on all training samples, on the clean (noisy) subset to obtain an \textit{expert} (\textit{anti-expert}) model. Finally, we adjust the parameters of base model by the difference between parameters of the \textit{expert} and \textit{anti-expert} models, steering the base model towards the  \textit{expert} model and away from the \textit{anti-expert} model. Experimental results show that CaPE improves performance across different automatic factual metrics and human evaluation, with the maximum improvement of 16.69\% and 15.78\% on summary-level dependency-arc entailment accuracy for the XSUM and CNN/DM datasets. The improvement in factual performance does not degrade the performance on other metrics of informativeness such as ROUGE. %\af{maybe metrics of informativeness such as ROUGE}%\footnote{Code is available at \url{https://github.com/salesforce/CaPE}}
\end{abstract}

\section{Introduction}
Neural abstractive summarization systems have been shown to generate plausible summaries with high lexical overlap with the references. 
However, human analyses \cite{fabbri2021summeval, pagnoni-etal-2021-understanding, tejaswin-etal-2021-well} and automatic evaluations \cite{falke-etal-2019-ranking, kryscinski-etal-2020-evaluating, maynez-etal-2020-faithfulness, durmus-etal-2020-feqa} show that state-of-the-art models trained on widely used XSUM \cite{narayan-etal-2018-dont} and CNN/DM \cite{DBLP:journals/corr/HermannKGEKSB15}  datasets tend to hallucinate information with high frequency. 
The degree of a model's hallucinations further correlate with the quality of training data \citep{aralikatte2021focus,pagnoni-etal-2021-understanding}.
For instance, models trained on the XSum data tend to generate a higher proportion of factual errors as compared to models trained on the CNN/DM dataset. %\jv{Maybe related to degree of hallucination in training data?}
%For instance, by performing human evaluations on 2250 model generated summaries from CNN/DM and XSUM datasets, \citet{pagnoni-etal-2021-understanding} found that 60\% of the summaries contained at least one factual error.  %\jw{bstractiveness and noise? citation here plz}. 
%
% The hallucinations are broadly classified as \textit{extrinsic}, when a model adds information that is not present in the source document, and \textit{intrinsic}, when the model distorts information present in the source document into a factually incorrect representation. 

\begin{table}[]
\small
\centering
\begin{tabular}{|c|cccc|}
\hline
Model & R-1 & R-2 & R-L & E-R$_{ref}$\\ \hline
All & 45.70 & 22.53 & 37.54 & 53.69 \\ 
Filtered & 41.66 & 18.39 & 33.66 & 42.58 \\ \hline
$\Delta$ & -8.84\% & -18.37\% & -10.33\% & -20.69\% \\
\hline
\end{tabular}%}
\caption{Validation performance comparison of BART models trained on all (204,017 samples) and filtered (50,270 samples) XSUM training data. %\jv{Since these are pretty large differences, it may be helpful to put absolute size of each dataset to better understand the context}
} \label{table-intro-result}
\end{table}

\begin{figure}[]
    \centering
    \includegraphics[width=\linewidth]{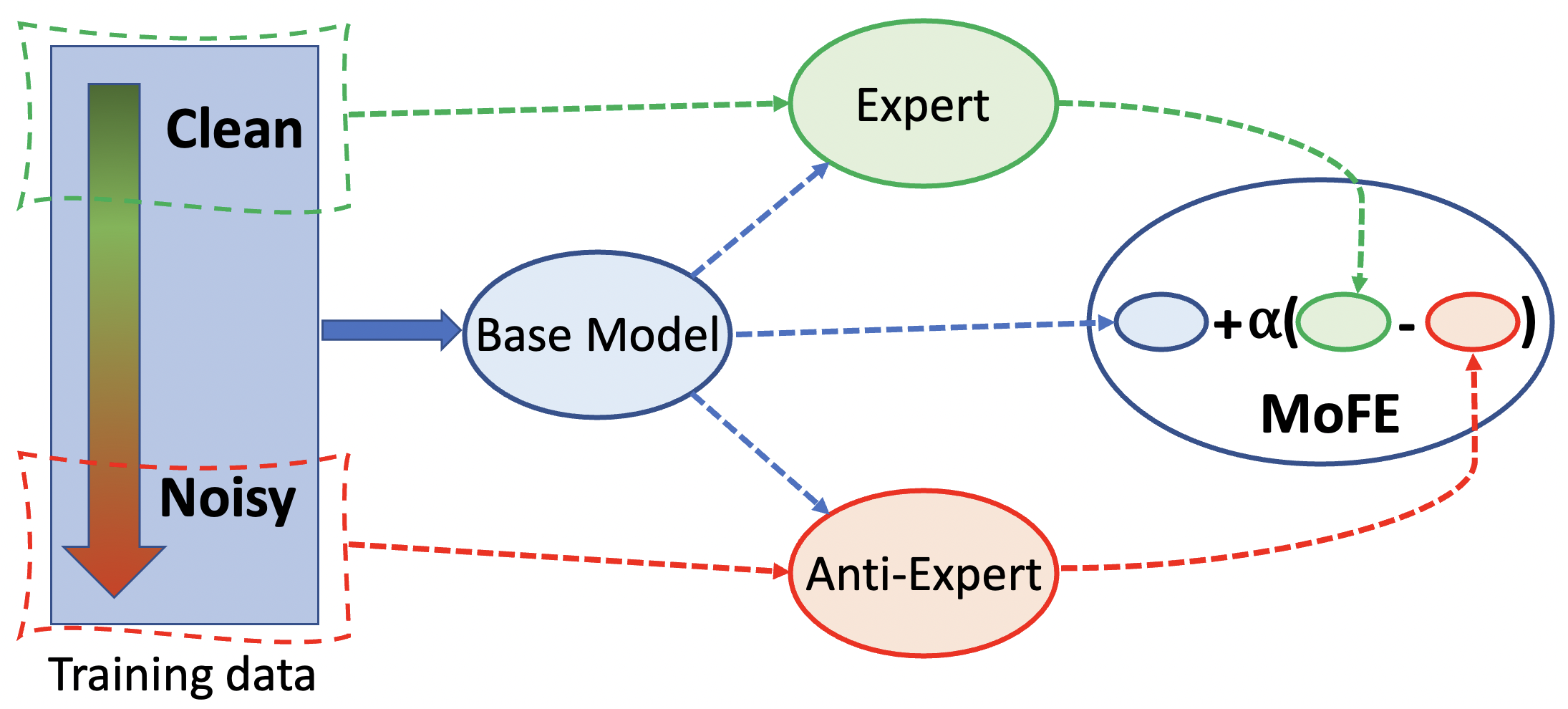}
    \caption{Schematic view of steps for building the CaPE model. First, it uses automated factual metrics to select clean and noisy training samples. Then, it fine-tunes an \textit{expert} and an \textit{anti-expert} on the clean and noisy training sets respectively, and uses them to readjust the parameters of the base summarization model. %caption to include details of the process
    } \label{model-flow}
\end{figure}

Given the association between training data quality and hallucinations in resulting models, the easiest method to reduce hallucinations is to remove noisy samples from the training data \citep{nan-etal-2021-entity}. %\jw{citation here}. 
However, data filtering reduces the size of training data and consequently the diversity in target summary since the removed noisy samples might also include useful task-specific knowledge. This impacts other aspects of generated summaries such as information recall or fluency. In Table \ref{table-intro-result}, we show ROUGE (R-1/2/L) and named entity recall (E-R$_{ref}$)  %\jw{need to be consistent, NER-RT or E-R$_{ref}$?} 
scores of a BART model \citep{lewis-etal-2020-bart} 
trained on the entity precision-filtered XSUM data (24.6\% of the original data). 
The new model drops 8-18\% in ROUGE and 20\% drop in entity recall. %\jw{ put this definition n the later section:(percentage of entities in the reference summary that are also present in the generated summary)}.
%\jw{ put this definition n the later section: filtered using entity overlap precision, i.e., the percentage of entities in summary that are present in the source document.} 
%\jw{cite BART here in the first time} 

In this work, 
we design a simple yet effective strategy to utilize both clean and noisy training samples. We use the observation that ``the level of hallucination in summarization model correlates with the level of noise in training data''.
Specifically, a model trained by maximizing the likelihood %(MLE) 
of a reference summary given its source document learns the hallucinations in training data. Therefore, we use an automatic factual metric to select clean data samples without any factual errors and fine-tune a base summarization model, which is trained on all data, to obtain an \textit{expert}. 
Similarly, we select noisy data samples that contain abundant factual errors and fine-tune the base summarization model to get an \textit{anti-expert}. The difference in factual qualities of data used to train our \textit{expert} and \textit{anti-expert} makes the \textit{anti-expert} hallucinate more than the \textit{expert}.
%\af{"filter" in the paper is mainly used as "select" whereas I would associate "filtering clean data samples" with removing those and leaving noisy ones, vice versa. Although it's usually clear which subset is being used, could swap filter with select or another word.} 
% we focus on the factual quality of training data and its association with the hallucinations in the summarization model. 

Next, we adjust base model's parameters by combining it with \textit{expert} and \textit{anti-expert}. Typically, when we have many models, a straightforward approach to combine them would be to take a weighted average of their output \citep{10.1162/neco.1991.3.1.79,liu-etal-2021-dexperts}. However, this requires running each model separately and increases computational cost linearly in the number of models, further slowing the auto-regressive generation of summaries. Alternatively, \citet{DBLP:journals/corr/abs-2001-01871} 
proposed attention over parameters that jointly optimizes multiple models and directly combines all their parameters through learned attention coefficients. 
% While taking the weighted average of parameters of independently trained models are unlikely to perform better than a randomly initialized model \citep{weight-ensemble,DBLP:journals/corr/abs-1912-05671,NEURIPS2020_0607f4c7}, recently,
Furthermore, 
\citet{wortsman2021robust} recently shows that average of a pre-trained CLIP \citep{DBLP:journals/corr/abs-2103-00020} model with its another version that is further fine-tuned on a new data distribution performs better than both models on their complementary distributions. Motivated by these findings and the fact that the \textit{anti-expert} possesses undesirable behavior, we propose Contrastive Parameter ensembling (CaPE),
a generalization of parameter averaging, which adds the \textit{expert}'s and subtracts the \textit{anti-expert}'s  parameters (equivalent to adding the difference between \textit{expert}'s and \textit{anti-expert}'s parameters) from the base model.

We evaluate our CaPE model on two benchmark abstractive summarization datasets, XSUM and CNN/DM. We train an \textit{expert} and an \textit{anti-expert} corresponding to each of the dependency-arc entailment %(D$_{arc}$) 
\citep{goyal-durrett-2020-evaluating,goyal-durrett-2021-annotating} and entity overlap %(E-P$_{src}$)
\citep{nan-etal-2021-entity} metrics. Then, we combine each \textit{expert} and \textit{anti-expert} pair
to obtain four variants of CaPE and evaluate them using the metrics used for data selection %D$_{arc}$/S, E-P$_{src}$, 
as well as a different entailment metric, MNLI  \citep{williams-etal-2018-broad}, and two question answering-based metrics, QuestEval \citep{scialom2021questeval} and QAFactEval \citep{DBLP:journals/corr/abs-2112-08542}, for factual consistency.  
% \jw{for evaluation or training or both, add some details here}. 
% We tried multiple combinations of experts and anti-experts with different splits, and 
We find that all variants of our CaPE consistently outperform the state-of-the-art models on all factual metrics, with marginal variations in ROUGE scores and information recall. 
% We find that all variants of CaPE models, obtained through different combinations of experts and anti-experts trained for entity precision and DAE metrics, strongly outperform the state-of-the-art models on factual consistency metrics, with marginal variations in ROUGE scores and recall. 
% \jw{the last sentence does not provide much information. Let's cover the number of improvement in what metric and key observation here.}
% Our empirical results suggest that we can steer the text summarization system to generate faithful content by effectively using clean and noisy data for training expert and anti-expert models respectively. 

% Our main contributions are:
% \begin{itemize}
%     \item We propose CaPE,
% \end{itemize}
\section{Contrastive Parameter Ensembling}
%
% \jv{What about keeping the terminology of ``Weight-space ensembling'' in order to tell a clean story about how you extend that existing approach? ``Parameter'' is a longer word and the phrase becomes a bit long overall. Also, wondering if there is a catchier acronmym here. ``C-WiSE'' is one possibility that keeps consistency with original name. Not sure how well the WiSE is known in the community}
% In this paper, we propose CaPE for improving factual consistency of text summarization systems. As illustrated in Figure \ref{model-flow}, CaPE mainly relies on training data selection for expert/anti-expert training and base model parameters re-adjustment based on expert and anti-expert parameters.
% \jv{
In this work, we propose Contrastive Parameter Ensembling (CaPE) for reducing hallucinations in text summarization systems. This method refines a base summarization model by training two additional models: an \textit{expert} model, which is trained on the subset of data with the highest factual consistency, and an \textit{anti-expert} model, trained on the subset of data with the lowest factual consistency. An ensemble model is then constructed through a simple linear combination of the parameters of the three models, an approach inspired by recent work on weight (a.k.a. parameter)-space ensembling~\cite{weight-ensemble,DBLP:journals/corr/abs-1912-05671,NEURIPS2020_0607f4c7,wortsman2021robust}.
% }

\subsection{Measuring Hallucinations for Selecting Training Data %\jv{Or, (Factual Consistency) Metrics for Selecting Training Data}
}
% \jv{Shall we talk about ``reducing hallucinations'' or ``improving factual consistency''? The latter frames it as a positive attribute that we're trying to improve, as opposed to a negative attribute that we're trying to reduce. I find the factual consistency framing slightly more intuitive, but don't feel strongly about it} 
To select data for training the \textit{expert} and \textit{anti-expert}, we assume the availability of  automated metrics for measuring hallucinations in reference summaries. There are several automatic metrics to evaluate factual consistency such as entity overlap \citep{nan-etal-2021-entity}, entailment score \citep{kryscinski-etal-2020-evaluating,goyal-durrett-2020-evaluating,maynez-etal-2020-faithfulness}, and QA-based metrics  \citep{durmus-etal-2020-feqa,scialom2021questeval}. %These methods have varied degree of computational cost \jv{
These methods vary greatly in computational cost and agreement with human judgements for factuality. We use the two of the faster metrics that are based on entity overlap and entailment metrics, and have shown good correlation with human-based evaluations, described below. 
% These methods measure different forms of hallucinations and vary greatly in computational cost. We use the two of the faster metrics that are based on entity overlap (extrinsic hallucinations) and entailment metrics (intrinsic hallucinations), described below.

% \jv{Maybe more direct like: we focus our analysis on entity overlap and entailment metrics because...something about QA metrics being slow}

\paragraph{Entity Overlap} is the simplest method measuring token-level overlap of the %information of interest, 
named entities, between the summary and source document \citep{nan-etal-2021-entity}. 
We use \textbf{entity token overlap precision} (E-P$_{src}$), %that is \jv{replace ``that is'' with comma maybe} 
the percentage of named-entities tokens in the summary that are also present in the source. This metric can be used as a proxy to measure simpler cases of hallucinations, such as out-of-article %\jv{out-of-article} 
entity errors \citep{pagnoni-etal-2021-understanding}, also known as %\jv{known as} 
\textit{extrinsic hallucinations} %\jv{<- italicize maybe} 
\cite{maynez-etal-2020-faithfulness}. % \jv{A human study, or human studies} 
A human study by \citet{pagnoni-etal-2021-understanding} finds 
this to be the most frequent form of error in models trained on XSUM data. However, it fails to capture intricate cases of hallucinations such as semantic frame errors (e.g., when an entity is present in the source but is attributed to the %\jv{a->the} 
wrong predicate). 
%\jv{Maybe justify focus on entities at the beginning of paragraph rather than stating they are the ``information of interest''. Perhaps mention extrinsic hallucinations specifically. }

\paragraph{DAE}
(Dependency Arc Entailment) measures fine-grained entailment by breaking the summary into smaller claims defined by dependency arcs, covering errors such as incorrect predicates or their arguments, coreference errors, discourse link errors, %etc. \jv{remove etc. is redundant with such as} 
in contrast to the simpler token-level entity overlap. Dependency arcs define grammatical structures in a sentence and often describe semantic connections between words, such as predicate-argument relations \citep{melcuk1988}. \citet{pagnoni-etal-2021-understanding} finds that DAE correlates with the human judgment of factuality, and has the highest correlation with complex discourse errors, such as entity coreference. Therefore, we use \textbf{DAE errors} (the number of dependency arcs in summary that are not entailed by the source document) to identify cases of more intricate hallucinations for selecting training data. % OOPS.. updated \jw{DAE errors --> ... dependency arcs in summary that are NOT entailed...?} % selection \jv{for selecting data or for the selection of training data}. 

% \jv{Wonder if it's worth drawing contrast to Entity overlap to illustrate they are complementary, e.g. ``Whereas Entity Overlap only captures extrinsic hallucinations, DAE also detects intrinsic hallucinations such as...''}

\subsection{Expert and Anti-expert based Model's Parameters Adjustment}

Using the entity overlap or DAE error metrics, we select samples for training \textit{expert} and \textit{anti-expert} models that %\jv{``which'' needs comma here, otherwise use ``that''} 
are then used to adjust the base model parameters. The data selection strategy, \textsc{SelectClean (SelectNoisy)}, and the generic process for building CaPE are described below and further illustrated in Algorithm \ref{alg:CaPE}. %\jv{Not sure if it is an issue that the notation is only introduced later. Maybe some rewording will help imply that more details (i.e. notation) are presented later.}

\paragraph{\textsc{SelectClean (SelectNoisy):}}
For the entity overlap metric, we select clean (noisy) samples with entity precision above (below) a predefined threshold $\epsilon^{E-P_{src}}_{clean}$ ($\epsilon^{E-P_{src}}_{noisy}$). For DAE error metric,  we select clean (noisy) samples with the number of DAE errors below (above) a predefined threshold $\epsilon^{DAE_{error}}_{clean}$ ($\epsilon^{DAE_{error}}_{noisy}$).

% \paragraph{\textsc{SelectNoisy:}}
% For the NER-P metric, we select noisy samples with entity precision below a predefined threshold $\Theta^{NER-P}_{noisy}$. Likewise, for DAE error metric,  we select noisy samples with the number of DAE errors above a predefined threshold $\Theta^{NER-P}_{noisy}$.

\paragraph{Fine-tuning Expert (Anti-expert)} %We fine-tune a base summarization model, trained on all training data, on the clean subset to obtain an \textit{expert} and on the noisy subset to obtain an \textit{anti-expert}. 
% \jv{
We train a base summarization model using all training data, and then fine-tune this model on the clean dataset to obtain the \textit{expert} and on the noisy dataset to obtain the \textit{anti-expert}.
% } 
% \jv{RAther than state as fact, state as aspiration, e.g. ``The goal of...''}
% T
By training on the full data followed by fine-tuning on \textit{clean} (\textit{noisy}) subset, we want our \textit{expert} (\textit{anti-expert}) model to retain other aspects such as ROUGE and information recall of the base model, and only differ in the factual qualities.  %\jv{Maybe stick with the parenthetical notation as before rather than slash} 
As noted in Table \ref{table-intro-result}, this is in contrast to training a BART model on just clean (or noisy) samples that severely deteriorates ROUGE and information recall (analyzed further in $\S$ \ref{ft-vs-rt}). 

% Secondly, training 
% an \textit{expert} model initialized with BART takes a  %training for 
% greater number of parameter updates ($>$ 1 epoch) to reach the best performance on ROUGE and other metrics. Contrarily, the base model already yields higher ROUGE score and fine-tuning it for 1 epoch is sufficient to reduce hallucinations, making fine-tuning a more efficient approach for building \textit{experts}  (\textit{anti-experts}). \af{Some of this text, like Secondly... feels like it could go later in analysis} %\jv{<- This could be clearer}

% \jv{Can you clarify? How is it fewer steps if also training on full dataset?}

Finally, for a mixing coefficient $\alpha$, we obtain our Contrastive Parameter Ensembled model ($\theta_{CaPE}$) from base ($\theta_B$), \textit{expert} ($\theta_E$) and \textit{anti-expert} ($\theta_{\bar{E}}$) parameters following: $\theta_{CaPE} = \theta_B + \alpha(\theta_E-\theta_{\bar{E}})$. %\jv{as follows: INSERT EQUATION} using the CaPE equation (described in the line \ref{eq:CaPE} of algorithm \ref{alg:CaPE}). 
The mixing coefficient ($\alpha$) balances factual quality with other aspects of summarization such as ROUGE and information recall.
% The mixing coefficient for expert and anti-expert models is used to control the factual quality and other aspects such as information recall of summaries generated by the CaPE model. \jv{The mixing coefficient balances factual quality with other qualities of summarization such as...}\jv{Give some hint here that it is a specified hyperparameter?}
%

\begin{algorithm}
\caption{CaPE for Summarization}\label{alg:CaPE}
\begin{algorithmic}[1]
\Require Training Data $D_T$, Measure of hallucination $M_H$
% \Ensure $y = x^n$
\State Train $\theta_B$ on $D_T$
\State $D_{clean}$ $\leftarrow$ \textsc{SelectClean} ($D_T$, $M_H$)
\State $D_{noisy}$ $\leftarrow$ \textsc{SelectNoisy} ($D_T$, $M_H$)
\State $\theta_E$ $\leftarrow$ Fine-tune $\theta_B$ on $D_{clean}$
\State $\theta_{\bar{E}}$ $\leftarrow$ Fine-tune $\theta_B$ on $D_{noisy}$
\State $\theta_{CaPE}$ $\leftarrow$ $\theta_B$ + $\alpha(\theta_E-\theta_{\bar{E}})$ \label{eq:CaPE}
\State \textbf{return} $\theta_{CaPE}$
\end{algorithmic}
\end{algorithm}
%

% CaPE comprising of either retrained BART or fine-tuned base summarization model-based \textit{expert} (\textit{anti-expert}) satisfies the condition of shared optimization trajectory that is generally required for parameter space ensembling. %Specifically, p
% \jv{
Initializing the \textit{expert} (\textit{anti-expert}) from the base or BART 
model is critical; prior work \citep{weight-ensemble,DBLP:journals/corr/abs-1912-05671,NEURIPS2020_0607f4c7} has shown that parameter-averaging works well when all %\jv{all -> the, since the past work only used 2 constituent models?} 
constituent models share the \textit{same} optimization trajectory. %}
% Prior work \citep{weight-ensemble,DBLP:journals/corr/abs-1912-05671,NEURIPS2020_0607f4c7} has shown that parameter-averaging works well when all constituent models share the optimization trajectory. 
On the other hand, averaging parameters of disjointly trained deep neural models, starting from different initializations, may not work better than a model with randomly assigned parameters. 
Since both methods of fine-tuning and training have a common initialization, % (either the base summarization model or the BART), 
the resulting CaPE model exhibits performance comparable to the base model or \textit{expert}.
%
% \begin{equation}
%     \theta_{CaPE} = \theta_B + \alpha(\theta_E-\theta_{\bar{E}}) 
%     \label{CaPE-eq}
% \end{equation}
%
%
%
\subsection{CaPE: A generalization of WiSE-FT}
% \vspace{1ex}
Contrastive Paremeter Ensembling generalizes the recently proposed WiSE-FT (Eq. \ref{wiseft-eq}) model \citep{wortsman2021robust}, which only performs a weighted sum of a base model and a single fine-tuned model,  
% Contrastive Parameter ensembling can be viewed as a \textit{generalization} of typical parameter averaging. \jv{It also seems like a special case, since it deals specifically with the contrastive learning. Maybe instead: Contrastive Weight Ensembling generalizes existing parameter averaging approaches, which only consider a weighted sum of a base model and a single fine-tuned model.} 
% Here, we compare our CaPE model ($\theta_{CaPE}$) with the
% WiSE-FT model , %a weighted average %space ensembling \jv{is there a more intuitive description than ``weight-space ensembing''? Not sure how widely known this is.} 
% of parameters of pre-trained and fine-tuned models, 
% proposed %by \citet{wortsman2021robust} 
for ensuring distributional robustness on image classification. 
\begin{equation}
    \theta_{WiSE-FT} = (1.-\alpha)\theta_B + (\alpha)\theta_E 
    \label{wiseft-eq}
\end{equation}
Essentially,  $\theta_{WiSE-FT}$ is a special case of $\theta_{CaPE}$ where the \textit{anti-expert} is a null (base) model.
% Comparing $\theta_{WiSE-FT}$ and $\theta_{CaPE}$, it is straightforward to conclude that WiSE-FT is equivalent to when \textit{anti-expert} model is replaced with the base model. \jv{<-It is also equivalent to the case where the anti-expert is essentially a null model  } 
We believe Eq. \ref{wiseft-eq} a sub-optimal solution for our objective of minimizing factual errors. Being trained on the noisiest subset of the training data, the \textit{anti-expert} model hallucinates with higher frequency than the base and \textit{expert} models, %.
% Thus, \textit{anti-expert} %provides greater contrast between the good and bad behavior than the base model, 
removing 
parameters responsible for hallucinations more than the other two. %the base model. 
We empirically find that our proposed contrastive ensembling outperforms the models that just use one of the \textit{expert} or \textit{anti-expert} in $\S$ \ref{expert-anti-expert}. %\jv{Wonder about factoring out and naming the idea of ``Contrastive Weight-Space ensembling'' (i.e. with expert and anti-expert) and introducing it as a novelty unto itself, apart from the application to factual consistency. This might be a candidate for part of the paper name, .e.g. ``Improving Factual Consistency through Contrastive Weight-Space Ensembling''}

\section{Results}
\subsection{Experimental Setup}
We evaluate CaPE on the XSUM \citep{narayan-etal-2018-dont} and CNN/DM \citep{DBLP:journals/corr/HermannKGEKSB15} datasets.
The XSUM data is highly abstractive and noisy. On the other hand, CNN/DM is more extractive and %but \jw{but? and?} 
contains fewer factual errors \citep{tejaswin-etal-2021-well}. These data variations allow us to evaluate CaPE under different data quality settings. 
Besides the standard ROUGE-1/2/L (R1/R2/RL) scores, we use a diverse set of metrics for evaluating factual consistency and summary quality. % of all models.

% \jw{I made it itemize here}
\begin{itemize}[leftmargin=*]
    \item \textbf{D$_{arc}$} measures the percentage of dependency arcs in summary that are entailed by the source article.
    
    \item \textbf{D$_{sum}$} measures the percentage of summaries that do not have any dependency arc error.

    \item \textbf{E-P$_{src}$} measures the percentage of entities in summary that are present in the source article.

    \item \textbf{E-R$_{ref}$} measures the percentage of entities in reference that are also present in the generated summary.

    \item \textbf{BS-P (R)} represents the BERTScore \cite{zhang2019bertscore} precision (recall) w.r.t. the source article.

    \item \textbf{QEval} represents a QA-based factual consistency metric ~\citep{scialom2021questeval}.
    
    \item \textbf{MNLI} measures the entailment score based on the RoBERTa large \citep{DBLP:journals/corr/abs-1907-11692} model trained on MNLI dataset \citep{williams-etal-2018-broad}. %\af{In case you want to add: 
    The score of a summary sentence is the maximum entailment score over all input sentences, and the final score is averaged across summary sentences as in \citet{10.1162/tacl_a_00453}.%}
    
    \item \textbf{QAFactEval} is another QA-based factual consistency metric that improves question filtering and answer overlap components  \citep{DBLP:journals/corr/abs-2112-08542}. %and achieves state-of-the-art performance on the SummaC benchmark \citep{10.1162/tacl_a_00453}. .
    
\end{itemize}

\begin{table*}[]
\small
\centering
\begin{tabular}{|l|cc|cc|c||cc|ccc||cc|}
\hline
Model & D$_{arc}$ & D$_{sum}$ & E-P$_{src}$ & E-R$_{ref}$ & QEval & BS-P & BS-R  & R1 & R2 & RL & TT & IT\\ \hline
\multicolumn{13}{|c|}{XSUM} \\ \hline
Base & 76.16 & 34.75 & 63.82 & {53.66} & 36.54 & 88.93 & 79.86 & \textbf{45.34} & {22.21} &  {37.13} & 1x & 1x\\
Ensemble &  75.22 & 33.48 & 62.63 & \textbf{54.23} & 36.37 & 88.82 & 79.86 & {45.27} & {22.28} & {37.09} & 1.2x & 1x \\ \hline
PP & 75.65 & 33.67 & 62.36 & 53.93 & 36.37 & 88.88 & 79.84 & 45.34 & \textbf{22.30} & 37.18 & 2-3x & 2x \\
PP-Clean & 79.41 & 40.09 & \textbf{72.98} & \underline{45.72} & 37.01 & 89.09 & 79.84 & 43.82 & 20.4 & 35.89 & 1.5x & 2x \\
PP-CC & 76.88 & 35.99 & 66.06 & 52.23 & 36.62 & 88.95 & 79.85 & 45.03 & 21.87 & 36.89 & - & 2x \\ \hline
CaPE$_{DD}$ & 78.48 & 39.14 & 65.52 & 53.0 & 36.90 & 89.06 & 79.83 & 45.32 & 22.26 & \textbf{37.22} & 1.07x & 1x \\ 
CaPE$_{PP}$ & 78.46 & 39.13 & {69.12} & 53.36 & 37.09 & 89.07 & \textbf{79.89} & 45.16 & 21.91 & 36.94 & 1.08x & 1x\\  
CaPE$_{DP}$ & \textbf{79.61} & \textbf{40.55} & 68.24 & 53.91 & \textbf{37.22} & \textbf{89.15} & \textbf{79.89} & 45.14 & 21.97 & 36.92 & 1.07x & 1x\\  
CaPE$_{PD}$ & 77.88 & 38.77 & 66.08 & 52.55 & 36.84 & 89.03 & 79.82 & 45.29 & 22.21 & 37.14 & 1.08x & 1x\\ \hline
CaPE$_{DP}*$ & \underline{83.87} & \underline{48.78} & \underline{74.3} & \underline{52.34} & \underline{38.05} & 89.41 & 79.93 & 43.56 & 20.39 & 35.46 & 1.07x & 1x\\  \hline
\multicolumn{13}{|c|}{CNN/DM} \\
\hline
Base &  96.26 & 75.0 & {98.44} & 58.92 & 59.24 & 93.26 & 82.62 & {44.05} & {21.07} & {40.86} & 1x & 1x\\
Ensemble & 95.19 & 67.44 & 97.72 & \textbf{61.93} & 59.51 &  93.06 & \textbf{82.91} & \textbf{44.28} & \textbf{21.23} & \textbf{40.88} & 1.2x & 1x\\   \hline
PP & 96.14 & 74.70 & 98.26 & 58.40 & 59.15 & 93.23 & 82.58 & 43.95 & 20.94 & 40.76 & 2-3x & 2x \\
PP-Clean & 96.17 & 74.77 & 98.63 & 58.20 & 59.16 & 93.23 & 82.59 & 43.92 & 20.92 & 40.74 & 2x & 2x \\
PP-CC & 95.72 & 72.63 & 98.52 & 58.57 & 59.11 & 93.22 & 82.61 & 43.97 & 20.98 & 40.79 & - & 2x \\ \hline
% Joint RL & \\
CaPE$_{DD}$ & \textbf{98.27} & \textbf{86.83} & 98.89 & 58.32 & \textbf{60.10} & \textbf{93.79} & 82.85 & 43.72 & 20.80 & 40.29 & 1.14x & 1x \\
CaPE$_{PP}$ & 97.17 & 80.46 & \textbf{99.16} & 58.66 & 59.65 & 93.52 & 82.71 & 43.62 & 20.72 & 40.33 & 1.14x & 1x\\
CaPE$_{DP}$ & 97.59 & 83.04 & 98.86 & {58.86} & 59.7 & 93.56 & 82.78 & 43.71 & 20.80 & 40.42 & 1.06x & 1x\\  
CaPE$_{PD}$ & 96.98 & 79.30 & 98.67 & 58.69 & 59.61 & 93.45 & 82.69 & 44.03 & 21.09 & 40.80 & 1.14x & 1x\\ \hline
% CaPE_{DP}* & \\  \hline
\end{tabular}
\caption{Performance comparison of CaPE and baseline models on XSUM and CNN/DM datasets. CaPE$_{DP*}$ is a variant of CaPE$_{DP}$ with $\alpha$  set  to 1.0. TT (IT) represents training (inference) time relative to the base model. %, D$_{arc}$: DAE arc-level accuracy, D$_{sum}$: DAE summary-level accuracy, E-P$_{src}$: entity precision w.r.t. source, E-R$_{ref}$: entity recall w.r.t. reference summary, QEval: QuestEval, BS-P/R: BERTScore precision/ recall, R1/2/L: Rouge 1/2/L. 
% \jw{only TT and IT were not introduced, others can be removed. And please add the definition of $CaPE_{DP*}$ here. Make sure all the columns are bold correctly, e.g., R2 PP in XSUM 22.30 should be bold instead of 22.28.} 
} \label{table-main-result}
\end{table*}

\subsection{Models}
We use the BART-based summarization (BART$_{sum}$) models released with Huggingface's transformers library \cite{wolf-etal-2020-transformers} (\textit{bart-xsum-large}, \textit{bart-cnn-large}) 
as the base models. 
From human-based analyses, \citet{pagnoni-etal-2021-understanding,fabbri2021summeval} find that BART$_{sum}$ models generated summaries have the least number of factual errors. %\af{although they didn't analyze Pegasus/T5. In SummEval we did see a very slight improvement of BART over Pegasus on CNNDM for factual consistency}
We adopt the standard hyper-parameters for all models during the inference.

% \jv{I feel like this notaton is a bit unintuitive} 
We train an \textit{expert} (\textit{anti-expert}) for each of the DAE error (Exp$_{DAE}$ (Anti$_{DAE}$)) and entity token overlap precision with source (Exp$_{E-P}$ (Anti$_{E-P}$)) metrics. %We construct two variants of CaPE, CaPE$_{W}$ and CaPE$_{L}$ using weights and logits ensembling respectively. 
% same and cross setting, alpha <1% reduction
We evaluate four variants of CaPE. CaPE$_{PP}$ uses Exp$_{E-P}$ and Anti$_{E-P}$, CaPE$_{DP}$ uses Exp$_{DAE}$ and Anti$_{E-P}$, and likewise. 
Depending on the value of $\alpha$, CaPE may reduce ROUGE or information recall while improving the factual consistency. Therefore, for each variant of CaPE, we select the $\alpha$ such that it does not under-perform the base model by more than 1\% on ROUGE 1 (R1) and entity recall (E-R$_{ref}$) metrics on the validation set.\footnote{%We set minimum value of $\alpha$ to 0.2 and increment it by the step size of 0.2.
We find $\alpha$ using grid search, assigning a minimum value of 0.2 and incrementing it by the step size of 0.2.}

\paragraph{Baselines:} We compare CaPE with two summarization baselines, BART$_{sum}$ (a.k.a. base) and an ensemble of BART-based summarization models, and three post-processing (PP) based models for improving factual consistency. Similar to CaPE, the ensemble model uses the average of a base summarization and two other summarization models obtained by fine-tuning the base model on two randomly sampled subsets of the training data.
For post-processing based models, we implement a variation of the autoregressive fact correction model from \citet{dong-etal-2020-multi}; we train a BART-large model to produce the reference summary conditioned on the concatenation of the source and reference summary with all entity slots masked. We call this model PP and train a variation of it on the subset of data with an entity precision of 100 (PP-clean). We also apply the model from \citet{chen-etal-2021-improving}, called PP-CC, that generates candidate summaries by enumerating all ways to replace entities in the summary with entities of similar type in the input and training BART with an additional classification layer to re-rank these summaries.

\begin{table}[]
\small
\resizebox{\columnwidth}{!}{
\centering
\begin{tabular}{|l|cc|cc|}
\hline
Model & \multicolumn{2}{c|}{XSUM} & \multicolumn{2}{c|}{CNN/DM}\\
& MNLI & QAFactEval & MNLI & QAFactEval \\ \hline
% Single & {25.77} & {36.54} & \textbf{38.22} &  {59.24}\\ \hline
Base & 22.70 & 2.104 & 84.20 & 4.550  \\
PP-Clean & 22.30 & 2.098 & 84.40 & 4.544  \\
CaPE$_{DP}$ & \textbf{23.10} & \textbf{2.205} & \textbf{86.80} & \textbf{4.602} \\  \hline
\end{tabular}}
\caption{MNLI and QAFactEval metrics-based evaluations of base, PP-clean and the CaPE$_{DP}$ model.} \label{table-qa-results}
\end{table}

% \subsection{Results}
\subsection{Automatic Evaluation} \label{automatic-evaluation}
Table \ref{table-main-result} summarizes the results on the XSUM and CNN/DM datasets.  %\textit{Ensemble$_{W}$} and \textit{ensemble$_{L}$} are weights- and logits-based ensemble of four BART models, including the best performing \textit{BART} (Single) model. 
First, we find that ensembling multiple %BART-based 
summarization models improves ROUGE scores, BERTScore recall and entity recall, but not necessarily factual consistency metrics. %defined by DAE accuracies and NER precision.
On the other hand, all variants of CaPE outperform the base %single BART 
as well as ensemble across all factual consistency metrics on both the XSUM and CNN/DM datasets. Given the controllability achieved by $\alpha$, we ensure that all variants of CaPE preserve ROUGE scores and information recall within a pre-defined threshold of maximum 1\% drop from the base model.
% On both XSUM and CNN/DM datasets, the worst being 0.55 drop compared to the best-performing ensemble$_L$ model on ROUGE-L for CaPE$_{L}$ on CNN/DM. 
We also find that CaPE models improve BERTScore precision (BS-P) with respect to the source article on both XSUM and CNN/DM. This is interesting given recent work on benchmarking different evaluation metrics that suggests that BERTScore precision with respect to the source document correlates with the human judgment of factuality \citep{pagnoni-etal-2021-understanding}.

% \paragraph{Comparison with post-processing based models for :}
CaPE models also outperform the post-processing based approaches PP and PP-CC on XSUM and all three PP, PP-clean and PP-CC approaches on CNN/DM dataset with significant margin. However, PP-clean performs similar to CaPEs on factual consistency metrics on XSUM and even obtains a higher E-P$_{src}$ score of 72.98. At the same time, PP-clean lowers the performance on ROUGE and information recall, reducing E-R$_{ref}$ performance by $\sim$15\%. Fortunately, we can %aggressively 
set the mixing coefficient $\alpha$ in CaPE to a higher value, achieving higher factual consistency at the cost of reduced ROUGE and information recall. To confirm this, we also report the performance of CaPE$_{DP}*$ on XSUM data which uses Exp$_{DAE}$ and Anti$_{E-P}$ mixed with  $\alpha$ value of 1.0 (underlined results in Table \ref{table-main-result}). We find that CaPE$_{DP}*$ obtains much higher score than PP-Clean model on all factual consistency metrics, while competently retaining the information recall of the base model (E-R$_{ref}$ reduced by 3.5\% compared to $\sim$15\% drop for PP-clean).

% In table \ref{table-main-result}, we evaluate our models using the three popular paradigms for evaluating factual consistency, namely, entity overlap, entailment and QA-based metrics.
% However, as noted by \citet{DBLP:journals/corr/abs-2112-08542}, prior studies on comparison of different factual metrics make inconsistent conclusions, with a few observing QA-based metrics superior to the entailment \citep{durmus-etal-2020-feqa,scialom2021questeval} while others reporting the opposite \cite{maynez-etal-2020-faithfulness}. Therefore, 

Finally, in Table \ref{table-qa-results}, we compare CaPE$_{DP}$ (the variant of CaPE with the best trade-off, discussed in $\S$\ref{alpha-tuning}), base and PP-clean models using two additional metrics, QAFactEval and MNLI. As noted by \citet{DBLP:journals/corr/abs-2112-08542}, prior studies comparing factual metrics draw inconsistent conclusions, with a few observing QA-based metrics as superior to entailment metrics \citep{durmus-etal-2020-feqa,scialom2021questeval} and others reporting the opposite \cite{maynez-etal-2020-faithfulness}. 
To the best of our knowledge, QAFactEval performs the best on the SummaC benchmark \citep{10.1162/tacl_a_00453}, used for comparing factual consistency metrics.
On both metrics, we find that CaPE$_{DP}$ outperforms both base and PP-clean models, %on both MNLI and QAFactEval score (scale of 1-5), 
improving the QAFactEval score by 4.8\% and 1.14\% over base model on XSUM and CNN/DM, respectively. 
% \vspace{-1ex}
% Across all metrics, we find CaPE models consistently reduce hallucinations as measured by all popular paradigms, namely, entity overlap, entailment and QA-based metrics, for evaluating the factual consistency.

\paragraph{Transferability of {Experts} ({Anti-experts}):}
We observe that CaPE models also improve performance on the metrics that were not used for training the \textit{expert} or \textit{anti-expert}. For instance, CaPE$_{PP}$ outperforms base model on the D$_{arc}$/D$_{sum}$ metrics, and CaPE$_{DD}$ outperforms base model on the E-P$_{src}$ metrics on both XSUM and CNN/DM. All variants of CaPE also outperform base model on QEval, QAFactEval and MNLI, % a question-answering based metric, 
which were also not used during the development of \textit{experts} (\textit{anti-experts}). 
Secondly, we find that the \textit{experts} and \textit{anti-experts} are interchangeable, an \textit{expert} trained on data selected using one metric can be used in conjunction with an \textit{anti-expert} based on another metric. As evident, both CaPE$_{DP}$ and CaPE$_{PD}$ outperform base model, with CaPE$_{DP}$ achieving best trade-offs among other variants of CaPE on the XSUM data, discussed further in $\S$\ref{alpha-tuning}.

% \vspace{-1ex}
\paragraph{Computational Efficiency:} We also report the approximate  training (TT) and inference (IT) time for different models relative to the base model in Table \ref{table-main-result}. We exclude the time required for data processing (e.g. data selection for CaPE and PP-Clean during training, or entity recognition for all post-processing based models both during training and inference). %Since the PP-CC model, 
We find that CaPE models only marginally increase the training time ($\leq$14\%) required for fine-tuning  \textit{expert} (\textit{anti-expert}) on a smaller selected subset of training data. Further, CaPE models do not increase the inference time. In comparison, post-processing methods use separate models for correcting summaries generated by the base model, increasing the memory required to store the additional model as well as both the training and inference time.

\subsection{Human Evaluation}
Following \citet{cao-wang-2021-cliff}, we also perform pairwise comparison of summaries, where human annotators rate each CaPE$_{DP}$ generated summary against the base model generated summary for factual consistency. %We use randomly sampled 100 articles from each of the XSUM and CNN/DM datasets.
% First, two annotators independently annotated 20 randomly sampled articles-summaries pairs from XSUM to calculate inter-annotator agreement. We found that two annotations achieve high Krippendoeff's alpha coefficient \cite{Krippendorff2011ComputingKA} of 0.83847. Then, one annotator 
We rate 100 random articles from each of the XSUM and CNN/DM datasets. The inter-annotator agreement is 0.8385 \cite{Krippendorff2011ComputingKA} based on our sampled  articles-summaries pairs from XSUM.
Annotators find CaPE$_{DP}$ improves (degrades) factual consistency on 19\% (14\%) summaries on XSUM data, and improves (degrades) factual consistency on 6\% (2\%) summaries on CNN/DM data. Factual consistency remained unchanged for the remaining 67\% and 92\% summaries from the XSUM and CNN/DM datasets, respectively.

\section{Analysis}

\subsection{Experts (Anti-experts) Performance}

\begin{table}[h]
\small
% \resizebox{\columnwidth}{!}
%{
\centering
\begin{tabular}{|l|cc|cc|c|}
\hline
Model & D$_{arc}$ & D$_{sum}$ & E-P$_{src}$ & E-R$_{ref}$ & R1\\ \hline
\multicolumn{6}{|c|}{XSUM} \\ \hline
Base & 76.16 & 34.75 & 63.82 & 53.66 & \textbf{45.34} \\ \hline
Exp$_{DAE}$ & \textbf{82.09} & \textbf{41.35} & 67.73 & 53.04 & 44.79  \\ 
Anti$_{DAE}$ & \underline{69.21} & \underline{17.52} & 58.63 & \textbf{56.95} & \underline{42.97} \\ \hline
Exp$_{E-P}$ & 78.81 & 36.42  & \textbf{69.81} & 51.60 & 44.53 \\ 
Anti$_{E-P}$ & 74.03 & 28.74 & \underline{57.15} & \underline{50.58} & 44.23 \\ \hline
\multicolumn{6}{|c|}{CNN/DM} \\ \hline
Base &  96.26 & 75.0 & \textbf{98.44} & \underline{58.92} & 44.05 \\\hline
Exp$_{DAE}$ & \textbf{97.50} & \textbf{80.40}&	98.30 &	60.42 & \underline{44.04} \\ 
Anti$_{DAE}$ & \underline{89.61} & \underline{42.75} & 96.69 & \textbf{62.14} & 44.07 \\ \hline
Exp$_{E-P}$ & 95.31 & 68.16 & 98.40 & 60.9 & \textbf{44.57} \\ 
Anti$_{E-P}$ & 93.48 & 57.85 & \underline{95.46} & 60.13 & 44.27 \\ \hline
\end{tabular}%}
\caption{Performance of individual \textit{experts} (\textit{anti-experts}) on the XSUM and CNN/DM datasets. Maximum scores are bolded and minimum scores are underlined for each of the metrics.} \label{table-expert-result}
\end{table}

In Table \ref{table-expert-result}, we compare the performance of individual \textit{expert} and \textit{anti-expert} models on DAE- and entity-based metrics. Our key findings include:

\begin{figure*}[]
     \centering
     \includegraphics[width=0.78\textwidth]{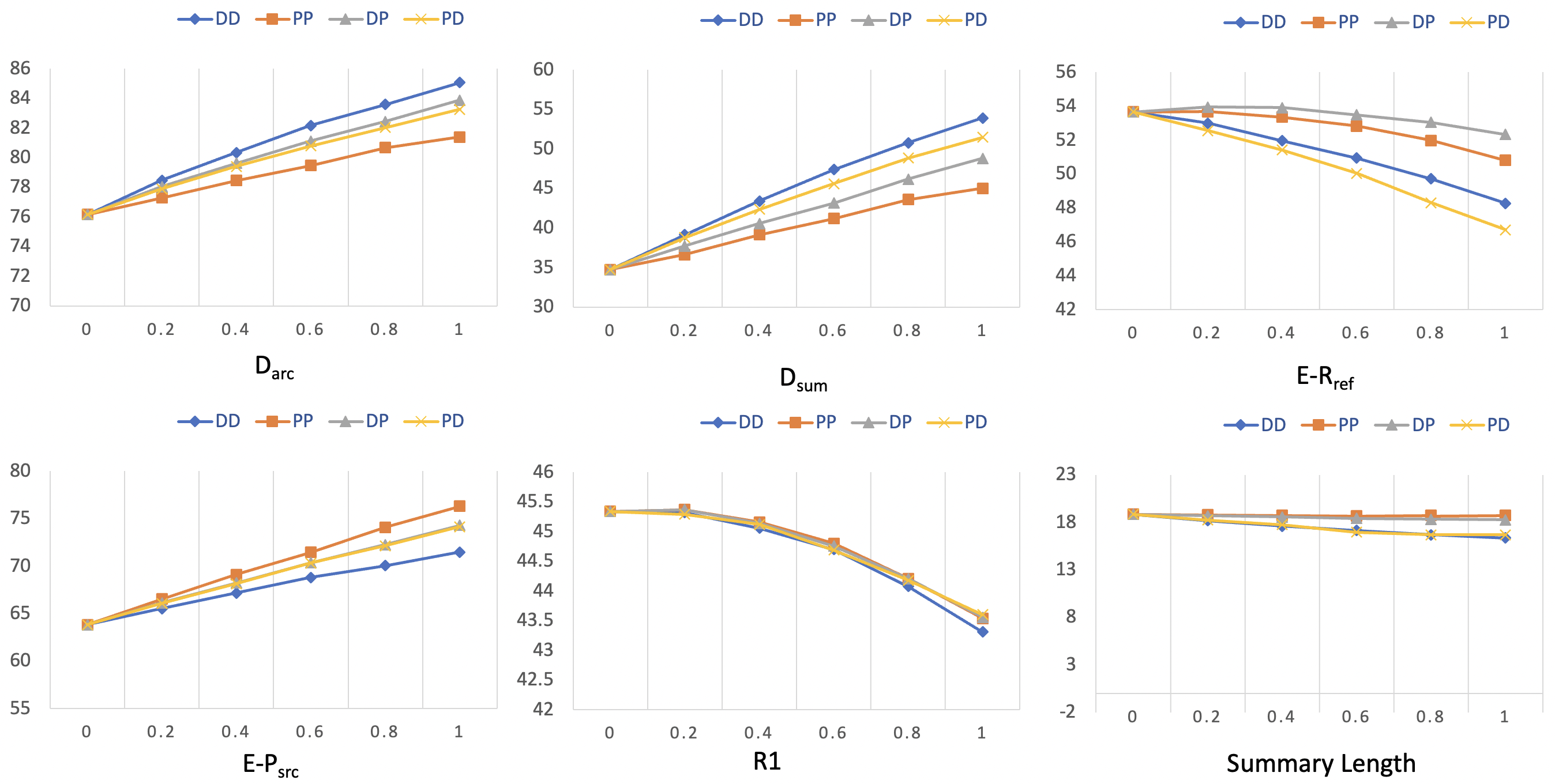}
     \caption{Variations in the performance of CaPE and base models with different values of mixing coefficient $\alpha$ on XSUM data ($\alpha$=0.0 corresponds to only base model.). }\label{fig:mixing-xsum}
\end{figure*}

\vspace{-1ex}
\paragraph{An expert  reduces hallucinations in generated summaries.}
We find that all \textit{experts}, except the entity-based \textit{expert} (Exp$_{E-P}$) on CNN/DM, are able to achieve improved performance on the metric used for selecting the training data subset. The unchanged performance of Exp$_{E-P}$ on CNN/DM is unsurprising given the base model is consistent against out-of-article entity error on CNN/DM dataset (E-P$_{src}$ of 98.44) and has very small room for improvement. This aligns with findings from human evaluation that the base model has very few extrinsic entity errors \citep{pagnoni-etal-2021-understanding}.
% \vspace{-1ex}
% \paragraph{An expert improves performance on the metric used for data filtering.}
On the noisy XSUM data, we observe that the improvement for \textit{experts} are not limited to the metrics used for data selection. For instance, Exp$_{DAE}$ improves entity precision (E-P$_{src}$) by $\sim$6\% and Exp$_{E-P}$ improves D$_{arc}$ and D$_{sum}$ by $\sim$3-4\%.

\vspace{-1ex}
\paragraph{An anti-expert increases hallucinations in generated summaries.}
All \textit{anti-experts} reduce performance on factual consistency metrics for both the XSUM and CNN datasets, with the maximum drop seen on summary-level D$_{sum}$ metric, indicating that a greater proportion of \textit{anti-expert} generated summaries are hallucinated. 
%While the lower performance for \textit{anti-experts} is a desirable behavior, we still require them to 
At the same time, they generate well-formed summaries, as indicated by their maintained ROUGE scores. This is the desirable behavior for an \textit{anti-expert} that should generate hallucinated but well-formed summaries.

\begin{table}[]
\small
\resizebox{\columnwidth}{!}{
\centering
\begin{tabular}{|l|c|cc|cc|}
\hline
Data & All & Exp$_{E-P}$ & Anti$_{E-P}$ & Exp$_{DAE}$ & Anti$_{DAE}$ \\ \hline
XSUM &  21.09 & 20.25 & 19.89 & 20.22 & \textbf{23.49} \\
CNN/DM & 51.57 & 48.4 &	51.07 &	\textbf{52.79} &	50.27  \\  \hline
\end{tabular}}
\caption{Average summary length of data used for training the base, \textit{expert} and \textit{anti-expert} models.} \label{table-summary-size}
\end{table}

\subsection{Effects of Mixing Coefficient $\alpha$} \label{alpha-tuning}
% \subsubsection{Effects of Mixing Coefficients on Ensemble of an Expert and BART }
% \begin{figure}[]
%      \centering
%      \includegraphics[width=\columnwidth]{dae.png} \\
%      \includegraphics[width=\columnwidth]{nerp.png}\\
%      \includegraphics[width=\columnwidth]{nerr.png}
%      \caption{Variations in the performance of weight-ensembled expert and BART models with different values of mixing coefficient $\alpha$.}\label{fig:mixing}
% \end{figure}

\begin{figure*}[]
     \centering
     \includegraphics[width=0.78\textwidth]{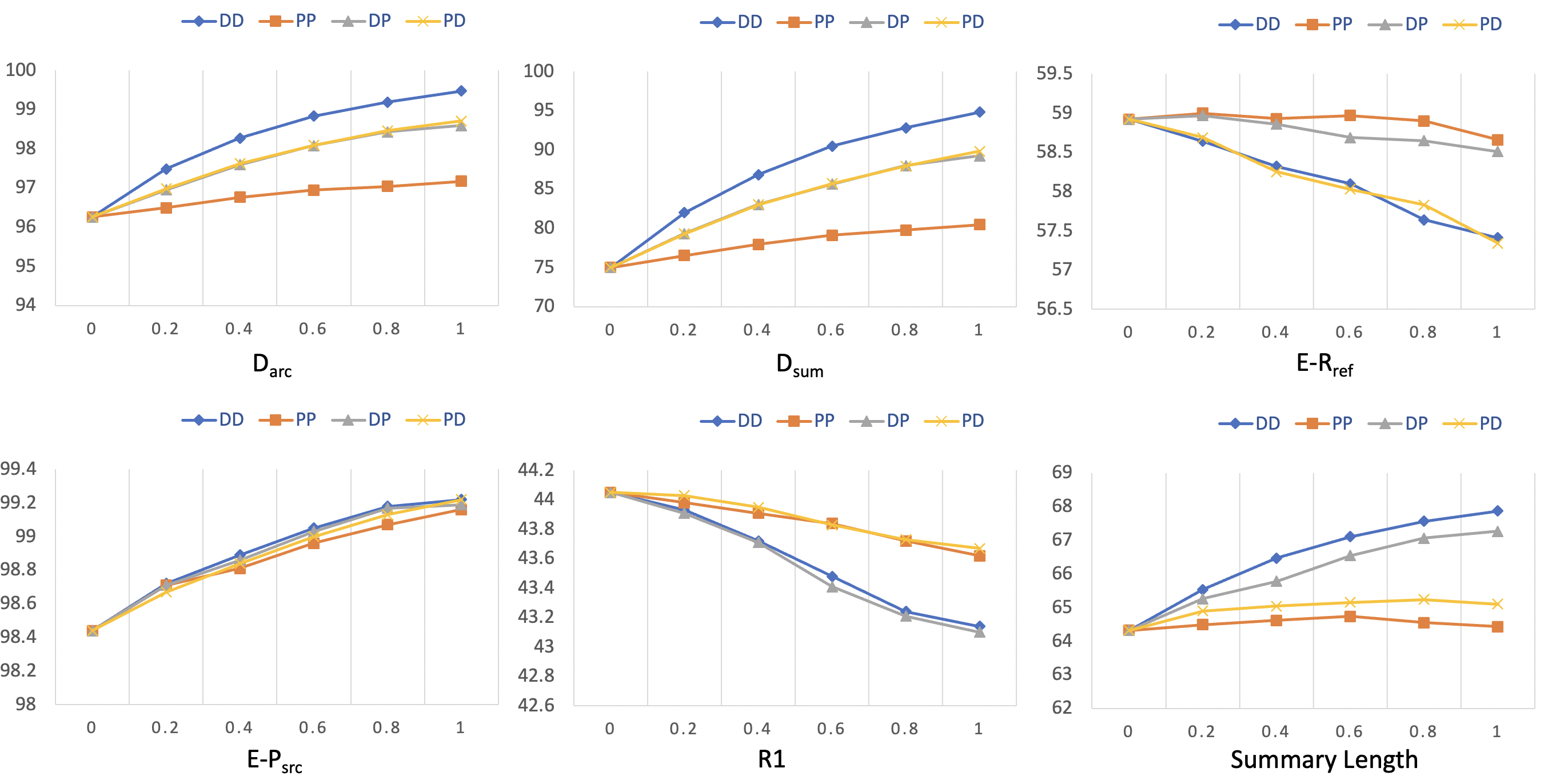}
     \caption{Variations in the performance of CaPE and base models with different values of mixing coefficient $\alpha$ on CNN/DM data ($\alpha$=0.0 corresponds to only base model.). }\label{fig:mixing-cnn}
\end{figure*}

We combine \textit{expert} and \textit{anti-expert} pair with the base model using different mixing coefficients ($\alpha$) and plot their performance on the XSUM and CNN/DM datasets 
in Figure \ref{fig:mixing-xsum} and \ref{fig:mixing-cnn}. We choose to vary $\alpha$ from 0.0 to 1.0. %\jw{if we dont have enough space, we can put CNN plot to the Appendix.}. 
We compare models on the D$_{arc}$/D$_{summ}$, E-P$_{src}$/RT, ROUGE 1 metrics. In addition, we compare the average summary length to capture artifacts introduced by data selection. We observe: 

% \jw{Why alpha  0 to 1.}
% \vspace{0.3ex}

\noindent \textbf{Inter-mixing the expert and anti-expert based on different metrics provides the best performance trade-offs.}
CaPE$_{DD}$, which uses the DAE-based \textit{expert} and \textit{anti-expert}, improves D$_{arc}$/D$_{summ}$ accuracy at the fastest rate on both datasets. Likewise, CaPE$_{PP}$ improves entity precision, E-P$_{src}$, at the fastest rate. CaPE$_{DP}$ and CaPE$_{PD}$ models that inter-mix the \textit{expert} and \textit{anti-expert} based on different metrics provide the best bargain on all factual consistency metrics, evenly improving all D$_{arc}$/D$_{sum}$ and E-P$_{src}$ scores.  
On the ROUGE score, we do not find any uniform pattern between the two datasets. 
On XSUM, all CaPE variants exhibit similar behavior while on CNN/DM, CaPEs using the entity precision-based \textit{anti-expert} (CaPE$_{PP/DP}$) retain ROUGE better than their alternatives.
Similarly, CaPE$_{PP/DP}$ retain entity recall better than their alternatives for all values of $\alpha$s on both datasets. Overall, CaPE$_{DP}$ provides the best balance for all performance measures on both datasets.

% \vspace{-0.05ex}
\noindent \textbf{Average summary length of data subset used for training {expert} ({anti-expert}) influences the length of CaPE-generated summaries.}
%Besides variations in performance measures, we observe an interesting association between the dataset and summary length. 
On XSUM data with shorter summaries, CaPE models tend to reduce the length of summary with increasing $\alpha$. Contrarily, on CNN/DM data with more extractive and longer summaries, models increase the average length of the summary with the increase in $\alpha$. From our initial analysis, as shown in Table \ref{table-summary-size}, this association can be explained by the average size of summaries in the data subset used for training \textit{expert} (\textit{anti-expert}). Specifically, CaPE$_{DD/DP}$ models see a maximum increase in the summary length on the CNN/DM dataset, which is confounded with the higher average summary length of data used for training the Exp$_{DAE}$ \textit{expert}. Similarly, on XSUM data, CaPE$_{DD/PD}$ models have a relatively lower average size than other models, which can be explained by the higher average summary length of samples used for training the Anti$_{DAE}$ \textit{anti-expert} (longer summaries for \textit{anti-expert} training makes CaPE generate shorter summaries). 

% Note that while CaPE leads to variations in the summary length, the difference is controllable through $\alpha$ and can be minimized by using lower values while also improving the factual consistency.

% On other metrics, ROUGE and entity recall, we find that the performance of the ensemble model either remains approximately unchanged (e.g. DAE-A, NER-PS metrics for the NER-R expert) or lies on the linear line (e.g. NER-PS/RT metrics for the DAE expert). Given the linear dependence, we can decide the mixing coefficient for an expert depending on the tolerance value for the ensemble model on all metrics. Further, we can
% compensate for the reduction in performance of the ensemble model on any metric by training an expert targeting that specific metric. For instance,  
% to compensate for the reduction in performance of the ensemble of DAE and BART on the NER-RT metric, we can add an NER-R expert that obtains higher NER recall than the base BART model.
% Note that, the modular characteristics of CaPE also allows us to choose different values of mixing coefficients for each of the experts and BART model depending on the significance of different factual errors in the target application.

\subsection{(Anti-)Expert Initialization: A Base Summarization Model outperforms BART} \label{ft-vs-rt}

\begin{figure}[]
     \centering
     \includegraphics[width=0.34\textwidth]{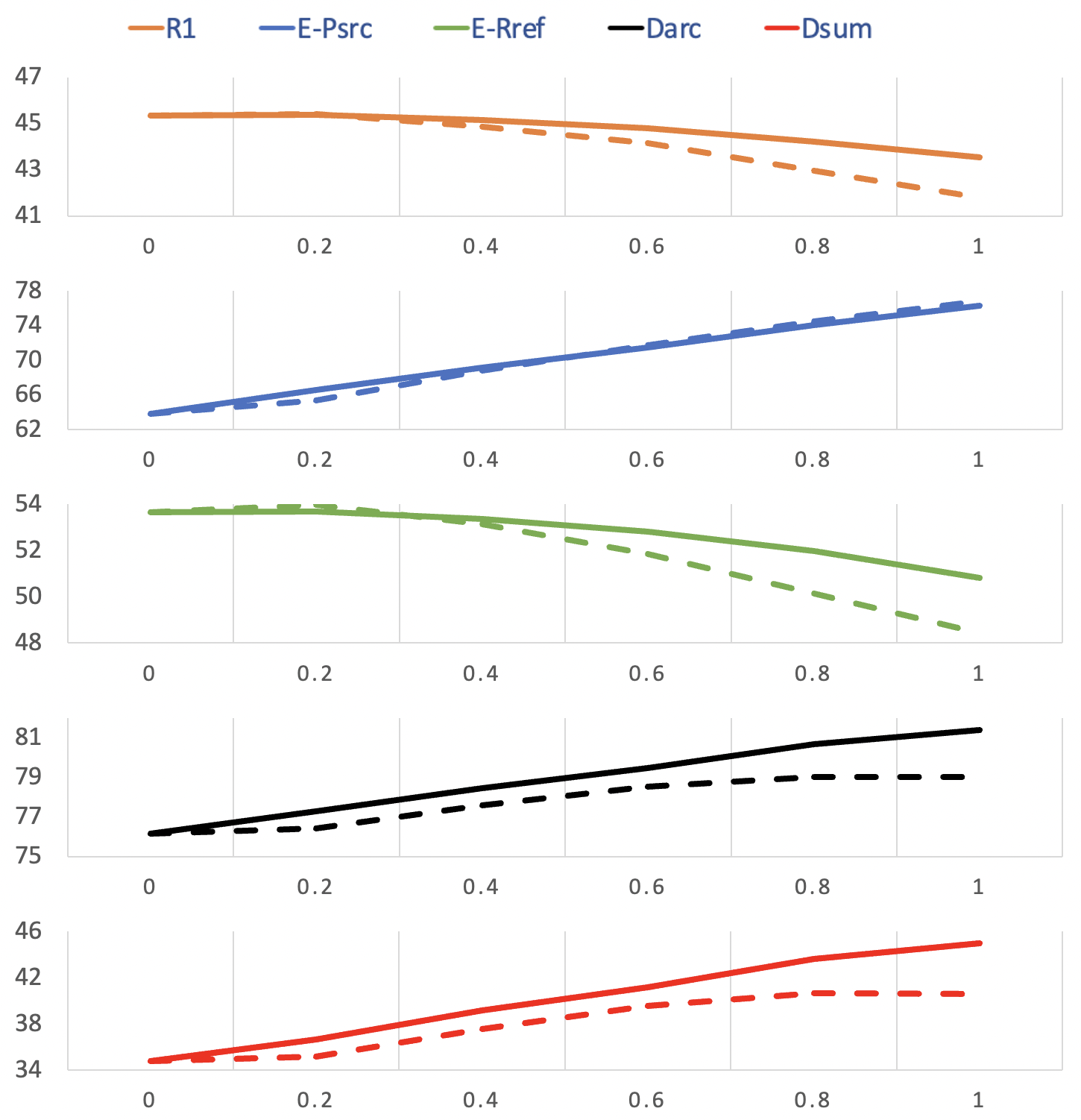}
     \caption{Performance comparison of models obtained by fine-tuning base summarization model (solid) vs training BART model (dashed) based on data selected according to the entity precision metric. %\jw{why the resolution is this low?} 
     }
     \label{fig:mixing-ft-rt}
\end{figure}

In Figure \ref{fig:mixing-ft-rt}, we compare the performance of CaPE$_{PP}$ models using \textit{expert} (\textit{anti-expert}) obtained by fine-tuning base summarization and training BART model. %\af{Previously (e.g. first sent in 4.2) BART model is synonymous with the base summarization model, so need to use different terminology, maybe pre-trained BART, but not sure.}
%\jw{I cant get what do you mean by finetuning and retraining here.}
First, we find that both models improve performance on all factual consistency metrics.
On the E-P$_{src}$ metric, which was also used to select the training samples, both models obtain comparable improvement. However, on the DAE-based factual consistency metrics as well as ROUGE and E-R$_{ref}$ metrics, fine-tuning the base model outperforms the one based on training BART. The gap in performance increases with the increase in value of $\alpha$, i.e., when the influence of \textit{expert} (\textit{anti-expert}) increases. This is unsurprising, given that the re-trained model leads to lower ROUGE and information recall (Table \ref{table-intro-result}) by being trained on fewer training samples.
Secondly, training 
an \textit{expert} model initialized with BART takes a  %training for 
greater number of parameter updates ($>$ 1 epoch) to reach the best performance on ROUGE and other metrics. Contrarily, the base model already yields higher ROUGE score and fine-tuning it for 1 epoch is sufficient to reduce hallucinations, making fine-tuning a more efficient approach for building \textit{experts}  (\textit{anti-experts}). %\af{Some of this text, like Secondly... feels like it could go later in analysis} %\jv{<- This could be clearer}

\subsection{CaPE outperforms Simple Parameter Ensembling (WiSE-FT)} \label{expert-anti-expert}

\begin{figure}[]
     \centering
     \includegraphics[width=0.34\textwidth]{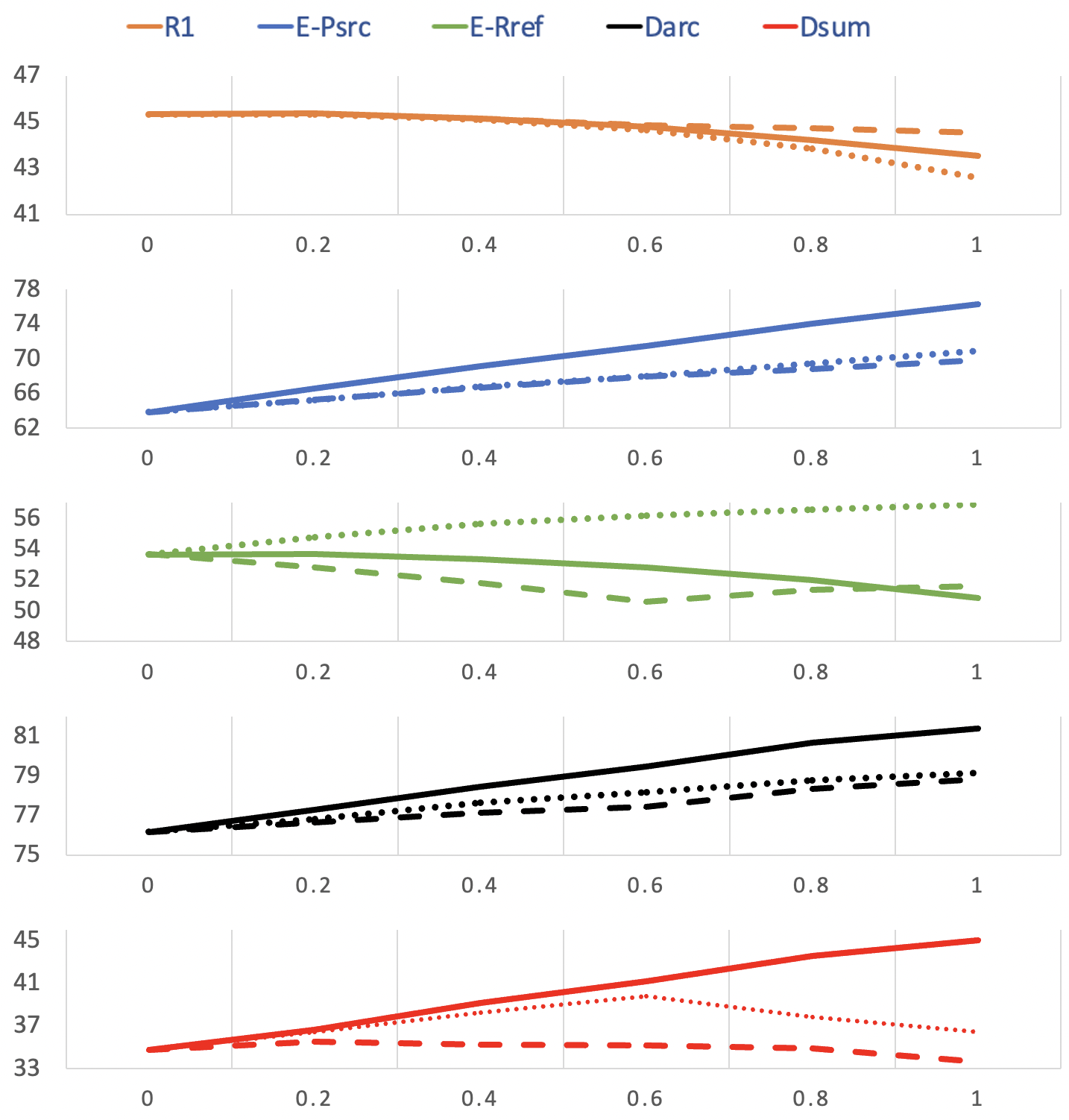}
     \caption{Performance comparison of CaPE (solid), \textit{Expert} only (dashed) and \textit{Anti-expert} only (dotted) models based on data selected according to the entity precision metric. }\label{fig:expert-anti-expert}
\end{figure}

In Figure \ref{fig:expert-anti-expert}, we compare the CaPE$_{PP}$ model with the \textit{expert} (\textit{anti-expert}) only model that replaces the \textit{anti-expert} (\textit{expert}) with the base model in $\theta_{CaPE}$. Accordingly, the \textit{expert} only model is equivalent to the WiSE-FT formulation ($\theta_{WiseFT}$). While both the \textit{expert} only and \textit{anti-expert} only improve performance on factual consistency metrics, we observe that CaPE$_{PP}$ improves performance %on factual consistency metrics 
at a faster rate than the former two models. %that use only one of the expert or anti-expert. 
On ROUGE-1 and E-R$_{ref}$ scores, the CaPE$_{PP}$ performance lies in between the \textit{expert} only and \textit{anti-expert} only models. The performance variations for the three models indicate that the contrastive ensembling combines the gains from \textit{expert} and \textit{anti-expert}, helping us to effectively use both clean and noisy data.

\section{Related Work}

% \paragraph{Factual consistency metrics and analysis}
Abstractive text summarization metrics such as ROUGE~\citep{lin-2004-rouge} and BERTScore~\citep{zhang2019bertscore} evaluate lexical and semantic overlap respectively but fail to sufficiently evaluate factuality and faithfulness~\citep{tejaswin-etal-2021-well}.
This has led to a line of research dedicated to evaluating factual consistency and hallucination in text summarization using new metrics such as entailment and question answering-based evaluation \citep{falke-etal-2019-ranking,kryscinski-etal-2020-evaluating, maynez-etal-2020-faithfulness,zhou-etal-2021-detecting,eyal-etal-2019-question,scialom-etal-2019-answers,wang-etal-2020-asking,durmus-etal-2020-feqa,scialom2021questeval}.
% All of these metrics can be used as proxies for entity-focused factual consistency evaluation and in particular QAGS, FEQA, and QuestEval have fact-based evaluation as their primary motivation. More recently, ~\citet{nan-etal-2021-entity} proposed entity-precision metric focusing on entity level factual consistency. 
% The slew of work on factual evaluation metrics has also given rise to research focused on comparing different metrics %, analyzing, and benchmarking 
% these metrics on various text summarization datasets. ~\citet{gabriel-etal-2021-go} show that although QA metrics are better than general metrics for evaluating factuality, they are extremely sensitive and there is no clear winner. 
% Some of the analysis work has focused on collecting human annotations for factual consistency errors, categorizing the errors, and measuring their correlations with automated metrics
% \cite{gabriel-etal-2021-go,fabbri2021summeval, pagnoni-etal-2021-understanding, goyal-durrett-2021-annotating, tejaswin-etal-2021-well}. 
%These evaluation studies often have contradicting observations. \af{suggestion: 
Research focused on comparing these factual consistency evaluation metrics \citep{gabriel-etal-2021-go,fabbri2021summeval, pagnoni-etal-2021-understanding, goyal-durrett-2021-annotating, tejaswin-etal-2021-well}, however, often have contradicting observations. %} 
For instance, \citet{durmus-etal-2020-feqa} found that entailment-based automated metrics have lower correlation with factual consistency  while \citet{pagnoni-etal-2021-understanding} concluded that the entailment-based FactCC %and semantic overlap-based BERTScore precision with respect to the source document 
exhibits the highest correlations with human judgments of factual consistency. %, and the correlation between FEQA and factual consistency is insignificant. 
Given the variations in findings from different human analyses of popular factual consistency evaluation metrics, we select a few metrics from each of the entailment, entity overlap, and QA-based evaluations, as well as use ROUGE and BERTScore metrics for evaluating CaPE.

% These works have created high quality datasets with annotations of different kinds of factual errors. AgreeSum~\citep{pang-etal-2021-agreesum} is another resource focusing on factuality applied to multi-document summarization.

% \begin{comment}
% \paragraph{Metrics and benchmarking metrics}
% \begin{enumerate}
%     \item FRANK ~\citep{pagnoni-etal-2021-understanding}
%     \item Go Figure ~\citep{gabriel-etal-2021-go}
%     \item SummEval~\citep{fabbri2021summeval}
%     \item FEQA ~\citep{durmus-etal-2020-feqa}
%     \item QAGS~\citep{wang-etal-2020-asking}
%     \item Entity level factual consistency metric and analysis~\citep{nan-etal-2021-entity}
    
% \end{enumerate}

% \paragraph{Datasets}
% \begin{enumerate}
%     \item AgreeSum ~\citep{pang-etal-2021-agreesum} -- analysis
%     \item Fine-grained factuality in cnndm/xsum ~\citep{goyal-durrett-2021-annotating} -- metrics
%     \item Analysis of summ datasets with factuality annotations~\citep{tejaswin-etal-2021-well} -- metrics/analysis
% \end{enumerate}

% \end{comment}

% \paragraph{Methods for enforcing factual consistency}
Along with the growing body of work on analysis and evaluation of factual consistency, there has been some recent work on developing methods to enforce factual consistency in pre-trained language models. These include sampling techniques such as constrained decoding~\citep{mao2020constrained} and neurologic decoding~\citep{lu2020neurologic}. 
Another strategy is to control generation either by using language models to guide a base language model as in GeDi~\citep{krause2020gedi} and DExperts~\citep{liu-etal-2021-dexperts} or via a hallucination knob~\citep{filippova-2020-controlled}. 
Although these methods claim to be generic, they have not been successfully applied to constrain summary generation on the source document. %Only very recently, ~\citet{xu2021dissecting} use the idea of dissecting generation modes to interpret the generated summaries.

Comparatively, there are fewer papers that propose methods for factual consistency in text summarization. Most of these focus on posthoc correction such as SpanFact~\citep{dong-etal-2020-multi}, contrast entity generation and selection~\citep{chen-etal-2021-improving}, loss truncation \cite{kang-hashimoto-2020-improved, goyal-durrett-2021-annotating}, and encoding SRL structure~\citep{cao-etal-2020-factual}.
~\citet{aralikatte2021focus} uses focus attention and sampling to improve the diversity and faithfulness of summaries while ~\citet{liu2021improving} uses data augmentation with a contrastive loss for factual consistency of abstractive summarization applied to customer feedback. 

Finally, works focusing on data noise include revising hallucinated summaries in training data \citep{https://doi.org/10.48550/arxiv.2204.10290}, dropping hallucinated samples (e.g. \citet{nan-etal-2021-entity} and \citet{narayan-etal-2021-planning} for summarization, \citet{matsumaru-etal-2020-improving} for headline generation), or defining curriculum based on the factual quality of training samples
\citep{kano-etal-2021-quantifying}.

% \af{If you want to add.- ~\citet{https://doi.org/10.48550/arxiv.2204.10290} propose a contrastive learning approach to revise hallucinations in reference summaries before training.}

% \af{Just a couple works which include filtering data as part of their factual consistency approach in case you want to add (one headline gen paper)}
% \cite{nan-etal-2021-entity, narayan-etal-2021-planning,matsumaru-etal-2020-improving}

% \begin{comment}
% \begin{enumerate}
%     \item Focus attention (FAME) for diversity and faithfulness~\citep{aralikatte2021focus}
%     \item \item Interpreting context vs. LM attribution~\citep{xu2021dissecting}
%     \item Hallucination knob by comparing to a LM ~\citep{filippova-2020-controlled}
%     \item Neurologic decoding~\citep{lu2020neurologic}
%     \item Constrained decoding~\citep{mao2020constrained}
%     \item DExperts~\citep{liu-etal-2021-dexperts}
%     \item GeDi~\citep{krause2020gedi}
%     \item Augmentation + constrastive loss~\citep{liu2021improving}
%     \item Posthoc contrast entity generation and selection~\citep{chen-etal-2021-improving}
%     \item Multifact correction (SpanFact)~\citep{dong-etal-2020-multi}
%     Encoding SRL structure~\citep{cao-etal-2020-factual}
    
% \end{enumerate}

% \paragraph{New citations:}
% \begin{enumerate}
%     \item method to control faithfulness-abstractiveness tradeoff~\citep{DBLP:journals/corr/abs-2108-02859,ladhak2021faithful}
%     \item factual consistency evaluation via masking~\citep{xie2021factual}
% \end{enumerate}
% \end{comment}

% \section{Factual Consistency Eva}

\section{Conclusion}
We present Contrastive Parameter Ensembling (CaPE) to reduce content hallucinations in abstractive summarization models. We first select clean (noisy) training samples to fine-tune an \textit{expert} (\textit{anti-expert}) model. Then, we use the difference between the parameters of \textit{expert} and \textit{anti-expert} models to adjust the parameters of a base summarization model. We evaluate CaPE on the XSUM and CNN/DM datasets using a diverse set of factual metrics, finding that CaPE effectively reduces hallucinations without a significant drop in ROUGE and information recall. %Further, our results and analyses highlight that text generation can be controlled for fine-grained factual qualities at decoding time through appropriately trained experts. % or increase in extractiveness. 

% \section{Ethics Statement}
% For crowD$_{sum}$ourced annotations, we did not collect any personal information from annotators and paid them above the minimum wage of \$15/hour. 

% Our method is evaluated on XSUM and CNN/DM datasets that contain English news articles crawled from western news outlets. While we expect our method to be adaptable to other languages, domains, or news articles from different geographies, we have not experimentally verified that.

\bibliography{custom}
\bibliographystyle{acl_natbib}

\clearpage
\appendix

\section{Experimental Details}
\paragraph{Data Selection:} For \textsc{SelectClean (SelectNoisy)}, we set $\epsilon^{E-P_{src}}_{clean}$,  $\epsilon^{DAE_{error}}_{clean}$, $\epsilon^{DAE_{error}}_{noisy}$ and  $\epsilon^{E-P_{src}}_{noisy}$ to 1.0, 0.0, 0.75 and 10.0 respectively.

\paragraph{Training Experts (Anti-experts):} We use Huggingface Transformers library \citep{wolf-etal-2020-transformers} (PyTorch \citep{paszke2017automatic}) to implement our \textit{experts} (\textit{anti-experts}). We initialize \textit{experts} with the pre-trained summarization models (\textit{bart-large-xsum}, \textit{bart-large-cnn}) and fine-tune them for 1 epoch with batch size of 64 using default training hyperparameters (optimizer: Adam, learning rate: 5e-5, ${\beta_1}$: 0.9, ${\beta_2}$: 0.999, ${\epsilon}$: 1e-8). The \textit{experts} (\textit{anti-experts}) initialized with BART are trained for 5 epochs. 

\paragraph{Inference:} We adopt the standard hyperparameters for all models during the inference, e.g. beam size of 6 (4), the minimum and maximum sequence length of 11 (56) and 62 (142), etc. for the XSUM (CNN-DM) model. %, and beam size of 4, minimum and maximum sequence length of 56 and 142, etc. for the CNN/DM model.

\end{document}